# Estimating Continuous Distributions in Bayesian Classifiers

George H. John
Computer Science Dept.
Stanford University
Stanford, CA 94305
gjohn@CS.Stanford.EDU
http://robotics.stanford.edu/~gjohn/

Pat Langley
Robotics Laboratory
Stanford University
Stanford, CA 94305
langley@CS.Stanford.EDU
http://robotics.stanford.edu/~langley/

## Abstract

When modeling a probability distribution with a Bayesian network, we are faced with the problem of how to handle continuous variables. Most previous work has either solved the problem by discretizing, or assumed that the data are generated by a single Gaussian. In this paper we abandon the normality assumption and instead use statistical methods for nonparametric density estimation. For a naive Bayesian classifier, we present experimental results on a variety of natural and artificial domains, comparing two methods of density estimation: assuming normality and modeling each conditional distribution with a single Gaussian; and using nonparametric kernel density estimation. We observe large reductions in error on several natural and artificial data sets, which suggests that kernel estimation is a useful tool for learning Bayesian models.

## 1 Introduction

In recent years, methods for inducing probabilistic descriptions from training data have emerged as a major alternative to more established approaches to machine learning, such as decision-tree induction and neural networks. For example, Cooper & Herskovits (1992) describe a greedy algorithm that determines the structure of a Bayesian inference network from data, while Heckerman, Geiger & Chickering (1994), Provan & Singh (1995), and others report advances on this basic approach. Bayesian networks provide a promising representation for machine learning for the same reasons they are useful in performance tasks such as diagnosis: they deal explicitly with issues of uncertainty and noise, which are central problems in any induction task.

However, some of the most impressive results to date have come from a much simpler -- and much older -- approach to probabilistic induction known as the *naive Bayesian classifier*. Despite the simplifying assumptions that underlie the naive Bayesian classifier, experiments on real-world data have repeatedly shown it to be competitive with much more sophisticated induction algorithms. For example, Clark & Niblett (1989) report naive Bayes producing accuracies comparable to those for rule-induction methods in medical domains, and Langley, Iba & Thompson (1992) found that it outperformed an algorithm for decision-tree induction in four out of five domains.

These impressive results have motivated some researchers to explore extensions of naive Bayes that lessen dependence on its assumptions but that retain its inherent simplicity and clear probabilistic semantics. Langley & Sage (1994) describe a variation that mitigates the independence assumption by eliminating predictive features that are correlated with others. Kononenko (1991) and Pazzani (1995) propose an alternative response to this assumption by selectively introducing combinations of attributes into the modeling process.

These and similar approaches represent an important line of research in machine learning, the goal of which is to discover learning methods that not only work well on real-world data but also have clear semantics. Although one means to this end is to study more modern systems and give a Bayesian interpretation, another research agenda begins with well-understood methods and attempts to improve on them by removing assumptions that might hinder performance.

In this paper, we take the latter approach, beginning with the naive Bayesian classifier, which traditionally makes the assumption that numeric attributes are generated by a single Gaussian distribution. Although a Gaussian may provide a reasonable approximation to many real-world distributions, it is certainly not always the best approximation. This suggests another direction in which we might profitably extend and improve the approach: by investigating more general methods for density estimation.

In the pages that follow we review NAIVE BAYES, the naive Bayesian classifier, then describe FLEXIBLE BAYES, an extension that eschews the single Gaus-sian assumption in favor of kernel density estimation (but which retains the independence assumption). We next discuss some important properties of kernel estimation that FLEXIBLE BAYES inherits. After this, we present some hypotheses about the new method's behavior, followed by experiments with natural and artificial domains designed to test those hypotheses. In closing, we review some related work and suggest some directions for future research.

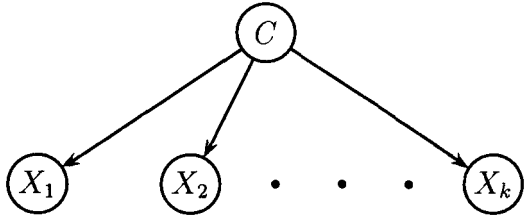

Figure 1: A naive Bayesian classifier depicted as a Bayesian network in which the predictive attributes $(X_1, X_2, \ldots X_k)$ are conditionally independent given the class attribute $(C)$.

## 2 The Naive Bayesian Classifier

As we have noted, the naive Bayesian classifier provides a simple approach, with clear semantics, to representing, using, and learning probabilitistic knowledge. The method is designed for use in supervised induction tasks, in which the performance goal is to accurately predict the class of test instances and in which the training instances include class information.

One can view such a classifier as a specialized form of Bayesian network, termed *naive* because it relies on two important simplifying assumptions. In particular, it assumes that the predictive attributes are conditionally independent given the class, and it posits that no hidden or latent attributes influence the prediction process. Thus, when depicted graphically, a naive Bayesian classifier has the form shown in Figure 1, in which all arcs are directed from the class attribute to the observable, predictive attributes (Buntine 1994).

These assumptions support very efficient algorithms for both classification and learning. To see this, let $C$ be the random variable denoting the class of an instance and let $\mathbf{X}$ be a vector of random variables denoting the observed attribute values. Further, let $c$ represent a particular class label, and let $\mathbf{x}$ represent a particular observed attribute value vector. Given a test case $\mathbf{x}$ to classify, one simply uses Bayes' rule to compute the probability of each class given the vector of observed values for the predictive attributes,

$$p(C = c | \mathbf{X} = \mathbf{x}) = \frac{p(C = c) p(\mathbf{X} = \mathbf{x} | C = c)}{p(\mathbf{X} = \mathbf{x})} \quad (1)$$

and then predicts the most probable class. Here $\mathbf{X} = \mathbf{x}$ represents the event that $X_1 = x_1 \wedge X_2 = x_2 \wedge \ldots X_k =$ $x_k$. Because the event is simply a conjunction of attribute value assignments, and because these attributes are assumed to be conditionally independent, one obtains

$$\begin{aligned} p(\mathbf{X} = \mathbf{x} | C = c) &= p(\bigwedge_i X_i = x_i | C = c) \\ &= \prod_i p(X_i = x_i | C = c) \quad, \end{aligned}$$

which is simple to compute for test cases and to estimate from training data. Generally one does not directly estimate the distribution in the denominator of Equation 1, as it is just a normalizing factor; instead one ignores the denominator and then normalizes so that the sum of $p(C = c | \mathbf{X} = \mathbf{x})$ over all classes is one.

NAIVE BAYES treats discrete and numeric attributes somewhat differently. For each discrete attribute, $p(X = x | C = c)$ is modeled by a single real number between 0 and 1 which represents the probability that the attribute $X$ will take on the particular value $x$ when the class is $c$. In contrast, each numeric attribute is modeled by some continuous probability distribution over the range of that attribute's values.

A common assumption, not intrinsic to the naive Bayesian approach but often made nevertheless, is that, within each class, the values of numeric attributes are normally distributed. One can represent such a distribution in terms of its mean and standard deviation, and one can efficiently compute the probability of an observed value from such estimates. For continuous attributes we can write

$$p(X = x | C = c) = g(x; \mu_c, \sigma_c) \text{ , where} \quad (2)$$
$$g(x; \mu, \sigma) = \frac{1}{\sqrt{2\pi}\sigma} e^{-\frac{(x-\mu)^2}{2\sigma^2}} \quad, \quad (3)$$

the probability density function for a normal (or Gaussian) distribution.[1]

The above model leaves us with a small set of parameters to estimate from training data. For each class and nominal attribute, one must estimate the probability that the attribute will take on each value in its domain, given the class. For each class and continuous attribute, one must estimate the mean and standard deviation of the attribute given the class. Maximum likelihood estimation of these parameters is straightforward. The estimated probability that a nominal random variable takes a certain value is equal to its sample frequency – the number of times the value was observed divided by the total number of observations. The maximum likelihood estimates of the mean and standard deviation of a normal distribution are the sample average and the sample standard deviation (Schalkoff 1992).

---

[1] Equation 2 is not strictly correct: the probability that a real-valued random variable exactly equals any value is zero. Instead we speak about the variable lying within some interval: $p(x \leq X \leq x + \Delta) = \int_x^{x+\Delta} g(x; \mu, \sigma) dx$. By the definition of a derivative, $\lim_{\Delta \to 0} p(x \leq X \leq x + \Delta)/\Delta = g(x; \mu, \sigma)$. Thus for some very small constant $\Delta$, $p(X = x) \approx g(x; \mu, \sigma) \times \Delta$. The factor $\Delta$ then appears in the numerator of Equation 1 for each class. They cancel out when we perform the normalization, so we may use Equation 2.

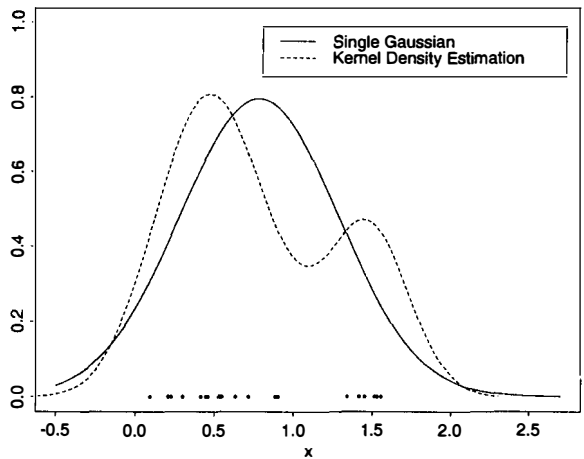

Figure 2: The effect of using a single Gaussian versus a kernel method to estimate the density of a continuous variable.

To clarify the estimation process, consider a small data set in which there are two classes (+ and −), a nominal attribute $X_1$ which takes values $a$ and $b$, and a continuous attribute $X_2$. Given the five training cases

$\{(+, a, 1), (+, b, 1.2), (+, a, 3.0), (-, b, 4.4), (-, b, 4.5)\}$,

NAIVE BAYES obtains the probability estimates

$$\begin{aligned} p(C = +) &= 3/5 \\ p(X_1 = a|C = +) &= 2/3 \\ p(X_1 = b|C = +) &= 1/3 \\ p(X_2 = x|C = +) &= g(x, 1.73, 1.21) \end{aligned}$$

for the positive class and analogous estimates for the negative class. The solid curve in Figure 2 shows the density function estimated for another numeric attribute from a somewhat larger data set.

One can also use Bayesian estimation methods, which commonly assume a Dirichlet prior, to estimate the model parameters. Despite the similar names, Bayesian estimation is less common in work on naive Bayesian classifiers, as there is usually much data and few parameters, so that the (typically weak) priors are quickly overwhelmed.

In summary, NAIVE BAYES provides a simple and efficient approach to the problem of induction. However, it typically relies on an assumption that numeric attributes obey a Gaussian distribution, which may not hold for some domains. This suggests that one should explore other methods for estimating continuous distributions.

Table 1: Algorithmic complexity for NAIVE BAYES and FLEXIBLE BAYES, given $n$ training cases and $k$ features.

|  | NAIVE BAYES | | FLEX BAYES | |
|---|---|---|---|---|
| Operation | Time | Space | Time | Space |
| Train on $n$ cases | $O(nk)$ | $O(k)$ | $O(nk)$ | $O(nk)$ |
| Test on $m$ cases | $O(mk)$ | | $O(mnk)$ | |

## 3   Flexible Naive Bayes

We can now introduce the FLEXIBLE BAYES learning algorithm, which is exactly the same as NAIVE BAYES in all respects but one: the method used for density estimation on continuous attributes. Although using a single Gaussian is the most common technique for handling continuous variables, it is certainly not the only one. Researchers have also explored a variety of nonparametric density estimation methods.

We chose to investigate *kernel density estimation*. Recall that in NAIVE BAYES we estimate the density of each continuous attribute as $p(X = x|C = c) = g(x, \mu_c, \sigma_c)$. Kernel estimation with Gaussian kernels (one can use other kernel functions as well) looks much the same, except that the estimated density is averaged over a large set of kernels

$$p(X = x|C = c) = \frac{1}{n} \sum_i g(x, \mu_i, \sigma_c) , \quad (4)$$

where $i$ ranges over the training points of attribute $X$ in class $c$, and $\mu_i = x_i$. The dashed line in figure 2 shows the kernel density estimation based on the 20 sampled points. Readers familiar with kernel methods should note that our Equation 4 is equivalent to the standard kernel density formula $p(X = x|C = c) = (nh)^{-1} \sum_j K(\frac{x-\mu_i}{h})$, where $h = \sigma$ and $K = g(x, \mu, 1)$.

Whereas in NAIVE BAYES one could estimate $\mu_c$ and $\sigma_c$ by storing only the sum of the observed $x$'s and the sum of their squares, the sufficient statistics for a normal distribution, FLEXIBLE BAYES must store every continuous attribute value it sees during training. The only sufficient statistic for the list of $\mu_i$'s is the list of $x_i$'s itself. (Nominal attributes' distributions are still learned by storing a single number per value that represents the sample frequency, as in NAIVE BAYES.) When computing $p(X = x|C = c)$ for a continuous attribute to classify an unseen test instance, NAIVE BAYES only had to evaluate $g$ once, but FLEXIBLE BAYES must perform $n$ evaluations, one per observed value of $X$ in class $c$. This leads to some increase in the storage and computational complexity, as summarized in Table 1.

We have not yet addressed the most important issue in kernel density estimation – the setting of the width parameter $\sigma$. As we shall see in the next section, kernel estimation has some nice theoretical properties when $\sigma$ shrinks to zero as the number of instances goes to infinity. The statistical literatures reports various rules of thumb for setting the kernel width, but each heuristic makes implicit and explicit assumptions about the density function that will be true of some distributions and not others. In this paper we set $\sigma_c = 1/\sqrt{n_c}$, where $n_c$ is the number of training instances observed with class $c$. Thus, as FLEXIBLE BAYES observes more training points, its density estimates become increasingly local.

The intuition behind FLEXIBLE BAYES is that kernel estimation will let the method perform well in domains that violate the normality assumption, with little cost in domains where it holds. To understand this claim, we must review the theoretical underpinnings of kernel methods.

## 4  Asymptotic properties of FLEXIBLE BAYES

In general, density estimation involves approximating the probability density function of a continuous random variable. The Bayesian classifier encounters this problem whenever it must estimate $p(X|C)$ for some continuous attribute $X$. This is a general problem in statistics, and a variety of methods are available for solving it (Venables & Ripley 1994, Silverman 1986). In this section we discuss the theoretical properties of kernel density estimation and their implications for the FLEXIBLE BAYES algorithm.

Statisticians are principally concerned with the *consistency* of a density estimate (Izenman 1991).

**Definition 1 (Strong Pointwise Consistency)** *If $f$ is a probability density function and $\hat{f}_n$ is an estimate of $f$ based on $n$ examples, then $\hat{f}_n$ is strongly pointwise consistent if $\hat{f} \to f(x)$ almost surely for all $x$; i.e., for every $\epsilon$, $p(\lim_{n\to\infty} |\hat{f}_n(x) - f(x)| < \epsilon) = 1$.*

The strongest asymptotic result we could hope for regarding FLEXIBLE BAYES would be that, provided the independence assumption holds, its estimate of $p(C|\mathbf{X})$ is strongly pointwise consistent. This would imply that, in the limit, using the FLEXIBLE BAYES estimate of $p(C|\mathbf{X})$ for classification produces the Bayes optimal error rate. We will prove strong consistency of FLEXIBLE BAYES in three steps, first proving that the method provides a strongly consistent estimate of $p(X|C)$ when $X$ is nominal, then proving the same property when $X$ is continuous, then proving that the estimate $p(C|\mathbf{X})$ is strongly consistent.

**Theorem 1 (Strong Consistency for Nominals)** *Let $X_1, \ldots, X_n$ be an independent sample from a multinomial distribution with $v$ values, where the probability of drawing value $j$ is $p_j$. Let $n_j = \sum_i 1_{X_i = j}$, the number of samples of value $j$. Then $n_j/n$ is a strongly consistent estimator of $p_j$.*

*Proof:* This is a direct instantiation of the strong law of large numbers (Casella & Berger 1990). ∎

**Theorem 2 (Strong Consistency for Reals)** *Due to Devroye (1983). The kernel density estimate is strongly consistent when:*

- *The kernel function $K$ must be a bona fide density estimate – it must be nonnegative for all $x$, and it must integrate to 1.*
- *$h_n \to 0$ as $n \to \infty$. Recall that $h$ in the standard notation is equivalent to our $\sigma$.*
- *$nh_n \to \infty$ as $n \to \infty$.*

All of these conditions are satisfied by using Gaussian kernels with $h_n = \sqrt{n}$, so each $p(X|C)$ density estimate in FLEXIBLE BAYES is strongly consistent.

**Lemma 1 (Consistency of Products)** *Let the functions $\hat{f}_1, \ldots, \hat{f}_k$ be strongly consistent estimates of density functions $f_1, \ldots, f_k$. Then $\prod_i \hat{f}_i$ is a strongly consistent estimator of $\prod_i f_i$.*

*Proof:* We prove that $\hat{f}_1 \hat{f}_2$ is a strongly consistent estimator of $f_1 f_2$, from which the lemma follows by induction. By the definition of strong consistency, for any $\epsilon_1$ and $\epsilon_2$, $p(\lim_{n\to\infty} |\hat{f}_{1,n} - f_1| < \epsilon_1) = 1$ for all $x$, and similarly for $\hat{f}_2$. For $\hat{f}_1 \hat{f}_2$ to be strongly pointwise consistent, $p(\lim_{n\to\infty} |\hat{f}_{1,n} \hat{f}_{2,n} - f_1 f_2| < \epsilon_3) = 1$ must hold for all $x$, and this will be true whenever $\epsilon_1 f_2 + \epsilon_2 f_1 + \epsilon_1 \epsilon_2 < \epsilon_3$ (by some simple algebraic manipulation). Since $\hat{f}_1(x)$ and $\hat{f}_2(x)$ are finite, and since $\epsilon_1$ and $\epsilon_2$ may be made arbitrarily small, the bound on $\epsilon_3$ can be made to hold, giving the desired result that $\hat{f}_1 \hat{f}_2$ is strongly consistent. ∎

**Theorem 3 (Consistency of FLEXIBLE BAYES)** *Let the true conditional distribution of the class given the attributes be $p(C|\mathbf{X}) = \dfrac{\prod_i p(X_i = x_i | C = c) p(C)}{\prod_i p(X_i = x_i)}$. (This is the actual conditional distribution, not our estimate.) Then the FLEXIBLE BAYES estimate $\hat{p}(C = c | \mathbf{X} = \mathbf{x})$ is a strongly consistent estimator of $p(C = c | \mathbf{X} = \mathbf{x})$.*

*Proof:* By Theorems 1 and 2, FLEXIBLE BAYES' estimates of $p(\mathbf{X}|C), p(X)$, and $p(C)$ are strongly consistent, thus by Lemma 1 and a related lemma regarding the quotient of strongly consistent estimates, FLEXIBLE BAYES' estimate of $p(C|X)$ is strongly consistent. ∎

Table 2: Natural data set characteristics and ten-fold cross validation results. Characteristics given are set size, number of classes, number of nominal and continuous attributes. Results given are the mean and standard deviations of the ten cross-validation runs for NAIVE BAYES and FLEXIBLE BAYES, along with the significance level of a paired $t$ test that one method is more accurate than the other. The accuracies for C4.5 are shown to provide context.

| Data set | Size | #Class | #Nom | #Cont | NAIVE | FLEX | FLEX Better? | | C4.5 |
|---|---|---|---|---|---|---|---|---|---|
| Breast Cancer (Wisc.) | 699 | 2 | 10 | 0 | 95.9± 0.2 | 96.7± 0.2 | ✓ | (99.0%) | 95.4 |
| Cleveland Heart Disease | 303 | 2 | 6 | 7 | 83.3± 0.6 | 80.0± 0.6 | × | (97.5%) | 72.3 |
| Credit Card Application | 690 | 2 | 6 | 9 | 74.8± 0.5 | 78.3± 0.6 | | | 85.9 |
| Glass | 214 | 7 | 9 | 0 | 42.9± 1.7 | 66.2± 0.9 | ✓ | (99.5%) | 65.4 |
| Glass2 (Float/Non) | 163 | 2 | 9 | 0 | 61.9± 1.4 | 83.8± 0.7 | ✓ | (99.5%) | 70.6 |
| Horse Colic | 368 | 2 | 7 | 15 | 73.3± 0.9 | 69.7± 1.0 | × | (99.5%) | 85.1 |
| Iris | 150 | 3 | 4 | 0 | 96.0± 0.3 | 95.3± 0.4 | | | 95.3 |
| Labor Negotiation | 57 | 2 | 8 | 16 | 86.0± 1.3 | 84.0± 1.6 | | | 85.7 |
| Meta-Learning | 528 | 2 | 19 | 3 | 67.1± 0.6 | 76.5± 0.5 | ✓ | (99.5%) | 72.6 |
| Pima Diabetes | 768 | 2 | 8 | 0 | 75.1± 0.6 | 73.9± 0.5 | | | 71.6 |
| Vehicle Silhouette | 846 | 4 | 18 | 0 | 44.9± 0.6 | 61.5± 0.4 | ✓ | (99.5%) | 70.0 |

## 5  Experimental Studies

However convincing our arguments for incorporating kernel estimation into the Bayesian classifier, the finite-sample behavior of this method is ultimately an empirical question. Within machine learning, the standard experimental method (Kibler & Langley 1988) involves running a learning algorithm on a set of training data, then using the induced model to make predictions about separate test cases and measuring the accuracy. To evaluate the behavior of the flexible Bayesian classifier, we designed and carried out a number of experimental studies along these lines.

### 5.1  Experiments on Natural Data

To determine the relevance of our approach to real-world problems, we first selected 11 databases from the UCI machine learning repository (Murphy & Aha 1994) and elsewhere. Table 2 summarizes the number of instances, the number of classes, and the number of nominal and numeric attributes in each data set. Because a sizeable fraction of each domain's features were numeric, they seemed likely candidates for contrasting the behavior of FLEXIBLE BAYES and NAIVE BAYES.

For each domain, we used ten-fold cross validation to evaluate the generalization accuracy of the two induction algorithms. That is, we randomly partitioned the data into ten disjoint sets, then provided each algorithm with nine of the sets as training data and used the remaining set as test cases. We repeated this process ten times using the different possible test sets and averaged the resulting accuracies. We also carried out this procedure with C4.5 (Quinlan 1993), a well-known algorithm for decision-tree induction, to provide a reference point for comparison.

Table 2 shows the results of the runs on natural domains, including the mean accuracy and standard deviation for each, whether one method was significantly better than the other on a paired $t$ test, and the significance level of that test. The table indicates that FLEXIBLE BAYES was significantly more accurate than NAIVE BAYES in five of the 11 domains, less accurate in two domains, and not significantly different in four cases. In two domains (Glass2 and Meta-Learning), the naive scheme was significantly worse than C4.5, whereas the flexible method did significantly better.

Curiously, the two domains where NAIVE BAYES outperformed FLEXIBLE BAYES were both medical domains. Possibly doctors tend to define diseases such that important continuous features are roughly normal, given whether a patient has a disease. If the conditional densities truly are Gaussian, NAIVE BAYES should perform better than FLEXIBLE BAYES (in the small-sample scenario) since the normality assumption is correct. To investigate further, we plotted histogram estimates of the density of each continuous attribute in the Cleveland heart disease dataset, conditioned on the class. On visual inspection, we did find that the distributions were roughly normal.

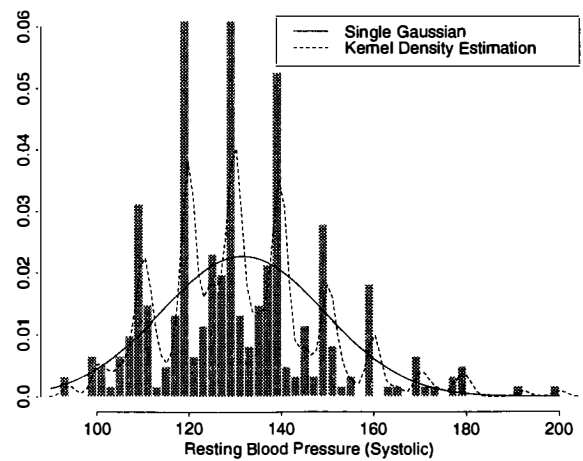

Figure 3: Systematic measurement errors in the Cleveland heart disease database.

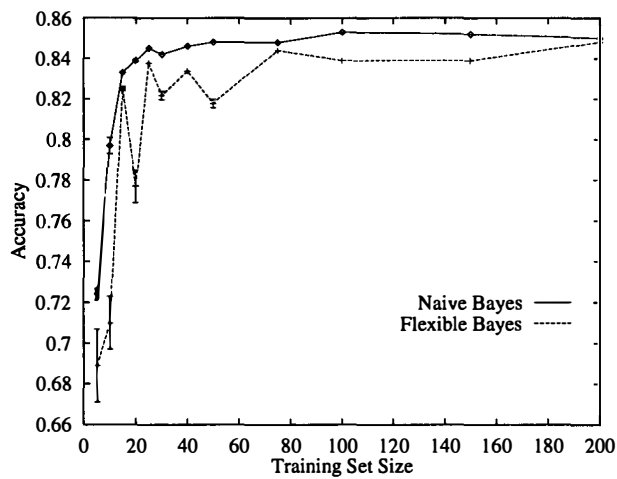
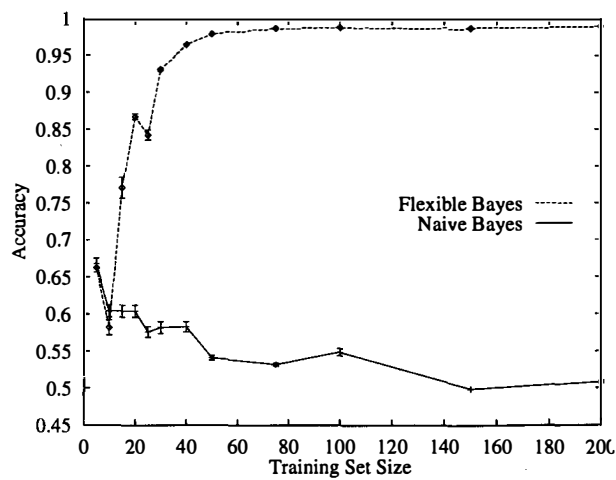

Figure 4: Learning curves testing the first hypothesis about behavior when the normality assumption holds.

Figure 5: Learning curves for the second hypothesis about behavior when normality is violated.

We were also quite surprised to discover systematic measurement errors. Figure 3 shows a surprising phenomenon working against the FLEXIBLE BAYES algorithm: the blood pressure of patients is sometimes rounded to the nearest 10 (note the peaks at 120, 130, 140), and sometimes not. There is no medical reason to suspect that a human's blood pressure is likely to be a multiple of ten, but when using the standard mercury-barometer/stethoscope apparatus for measuring blood pressure, it is difficult to get more than two significant figures, so rounding is common (Tigrani 1995, personal communication). The smooth curve shows the Gaussian density estimate, while the rough curve shows the kernel density estimate, which is confused by the rounding errors in recording the blood pressure of patients. We hypothesized that if this attribute were removed, FLEXIBLE BAYES and NAIVE BAYES would perform comparably. We ran a ten-fold cross-validation experiment with the blood pressure attribute removed, and found that both methods achieved an accuracy rate of 84.66%.

### 5.2 Artificial Domains

The positive results on the natural domains give us confidence in the usefulness of FLEXIBLE BAYES as a tool for machine learning, but since the real domains are themselves poorly understood, scientifically they do not let us draw any hard conclusions. Several of our expectations could not be directly tested on the natural databases. These expectations were:

1. When the normality assumption holds (when each $p(X|C)$ is normal), FLEXIBLE BAYES and NAIVE BAYES should exhibit the same asymptotic performance, but NAIVE BAYES should reach asymptote faster (with smaller training sets).
2. When the independence assumption holds but the normality assumption does not, FLEXIBLE BAYES should reach the Bayes optimal error rate while NAIVE BAYES should not.

To test these hypotheses, we defined appropriate joint probability distributions and then independently sampled varying-sized training sets from these distributions in order to plot learning curves for the two algorithms. We constructed a test set containing one thousand instances sampled from the same distribution, which we used to evaluate the models induced from the training sets. For each training set size, ten independent samples were constructed, and we report the average and standard deviation for the ten runs.

To examine the first hypothesis, we defined a joint probability distribution over one continuous attribute and one binary class, such that the conditional distribution of the attribute given the class was Gaussian. The learning curves in Figure 4 support our hypothesis: although FLEXIBLE BAYES performs worse than NAIVE BAYES when the size of the training set is small, its performance approaches that of NAIVE BAYES as the training set grows.

To test the second hypothesis, we defined a joint probability distribution over one continuous attribute and one binary class, such that the conditional distribution of the attribute given the class was a mixture of two Gaussians. The learning curves in Figure 5 again support our hypothesis. NAIVE BAYES is helpless because it is learning with an overly parsimonious model, while FLEXIBLE BAYES' flexibility lets it fit this distribution as well. With a training set of 200 instances, FLEXIBLE BAYES' accuracy was 99.0%. Using numerical integration, we estimated the Bayes optimal error rate in this domain to be 1.863%. The apparent impossibility of the situation (namely, FLEXIBLE BAYES outperforming the Bayes optimal error rate) is explained by realizing that the estimate of the algorithm's performance comes from a test set of 1000 instances. We expect that, had we used an order of magnitude more test instances, the algorithm's error would match the Bayes optimal error.

In summary, we have compared our FLEXIBLE BAYES algorithm with its natural benchmark, NAIVE BAYES, on a variety of natural and artificial domains, finding encouragement from the former and evidence for our hypotheses in the latter.

## 6 Discussion

Although our approach to Bayesian induction is novel, it does bear some similarities to other research in machine learning and statistics. The use of density estimation figures prominently in several learning algorithms. Specht & Romsdahl (1994) is the latest in a series of papers on kernel estimation in the guise of "probabilistic neural networks." In contrast to our approach, their method only handles continuous features, and makes no independence assumptions, so that a single $d$-dimensional density estimation is done per class, rather than $d$ one-dimensional estimations. A novel feature is their use of the conjugate gradient algorithm to optimize their cross-validation estimate of error over the space of smoothing parameters (one parameter per dimension). This gave tremendous improvement on some domains, and we expect that an adaptive kernel width would improve our results as well.

A histogram is one of the oldest and simplest methods of density estimation. Kononenko (1993) reports the use of experts to discretize continuous features. In contrast, Dougherty, Kohavi & Sahami (1995) study several methods for automatically discretizing continuous features based on statistical and information-theoretic metrics. They report that a naive Bayesian classifier combined with discretization gives higher accuracy than C4.5, averaged over many domains.

The algorithms explored in this paper are but two samples from a large space of possible algorithms that our framework suggests. Figure 6 gives a perspective on the work discussed in this paper, showing how NAIVE BAYES and FLEXIBLE BAYES relate to each other and to previous work.

The representational power of the flexible naive Bayesian classifier seems quite similar to that for Generalized Additive Models of Hastie & Tibshirani (1990) for regression. Their method predicts the value of a continuous variable, given various nominal and numeric input variables, using

$$\hat{y} = f_0(\sum_i f_i(x_i)) \ ,$$

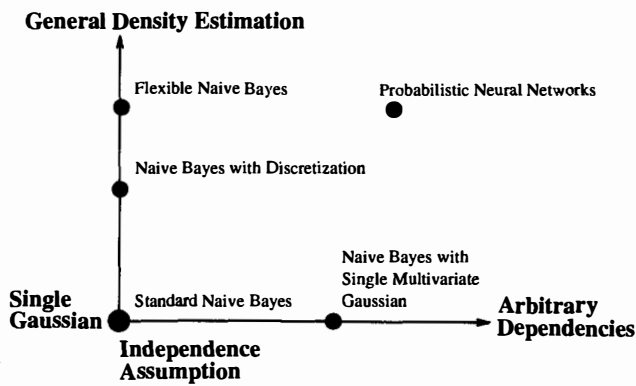

Figure 6: The space encompassing the algorithms discussed in this paper.

where the $f_i$ are arbitrary (possibly nonlinear) functions. By taking logarithms, Equation 1 may be transformed into a generalized additive model.

Two approaches seem close to our method at first glance, but upon close inspection there are important differences. Geiger & Heckerman (1994) present a method for learning "Gaussian networks", which are Bayes nets with some continuous variables estimated by a single Gaussian distribution, in contrast to our approach in which densities are estimated kernel estimation. Kononenko (1993) uses discretization with a "fuzzy" modification. Let $x$ be the value of a continuous feature in a test instance. Rather than assigning $x$ to a single interval, Kononenko "fuzzifies" $x$ using a Gaussian and assigns probabilistically to several intervals. The use of the Gaussian at first seems similar, but in our work such kernels are used only to obtain a smooth density estimate on the training data.

In future research, we hope to extend FLEXIBLE BAYES to set the kernel width adaptively. Cross-validation and other resampling schemes are relatively cheap to use in this context. For the implementation discussed here, the complexity of cross-validation is $O(n^2k)$, so a wrapper method for setting $\sigma$ similar to that reported by John, Kohavi & Pfleger (1994) may be employed on small- to medium-sized databases. A further possibility is borrow Cheeseman et al.'s (1988) Bayesian approach to density estimation with Gaussian mixtures using the EM algorithm (Dempster, Laird & Rubin 1977).

## 7 Conclusion

In this paper we reviewed the naive Bayesian classifier and the assumptions on which it relies, including the common use of a single Gaussian distribution for each predictive attribute. We argued that this assumption might be violated in some domains, and we proposed instead to use a kernel estimation method to approximate more complex distributions. Experiments with natural domains showed that, in a number of cases, this flexible Bayesian classifier generalizes better than the version that assumes a single Gaussian. Experimental studies with artificial data further suggest that the approach behaves as expected in two important scenarios. Although more work remains to be done, our results to date indicate that the flexible Bayesian classifier constitutes a promising addition to the repertoire of probabilistic induction algorithms.

## Acknowledgments

The authors would like to thank Nils Nilsson and Wray Buntine for their helpful comments and pointers to relevant literature. George John was supported by an NSF Graduate Research Fellowship, and Pat Langley was supported in part by Grant No. N00014-94-1-0746 from the Office of Naval Research.